# Vision-Based Assessment of Parkinsonism and Levodopa-Induced Dyskinesia with Deep Learning Pose Estimation

Michael H. Li, Tiago A. Mestre, Susan H. Fox, Babak Taati*

*Abstract— Objective:* To apply deep learning pose estimation algorithms for vision-based assessment of parkinsonism and levodopa-induced dyskinesia (LID). *Methods:* Nine participants with Parkinson's disease (PD) and LID completed a levodopa infusion protocol, where symptoms were assessed at regular intervals using the Unified Dyskinesia Rating Scale (UDysRS) and Unified Parkinson's Disease Rating Scale (UPDRS). A state-of-the-art deep learning pose estimation method was used to extract movement trajectories from videos of PD assessments. Features of the movement trajectories were used to detect and estimate the severity of parkinsonism and LID using random forest. Communication and drinking tasks were used to assess LID, while leg agility and toe tapping tasks were used to assess parkinsonism. Feature sets from tasks were also combined to predict total UDysRS and UPDRS Part III scores. *Results:* For LID, the communication task yielded the best results for dyskinesia (severity estimation: $r$ = 0.661, detection: AUC = 0.930). For parkinsonism, leg agility had better results for severity estimation ($r$ = 0.618), while toe tapping was better for detection (AUC = 0.773). UDysRS and UPDRS Part III scores were predicted with $r$ = 0.741 and 0.530, respectively. *Conclusion:* This paper presents the first application of deep learning for vision-based assessment of parkinsonism and LID and demonstrates promising performance for the future translation of deep learning to PD clinical practices. *Significance:* The proposed system provides insight into the potential of computer vision and deep learning for clinical application in PD.

Research supported by the Natural Sciences and Engineering Research Council of Canada (NSERC), the Toronto Rehabilitation Institute–University Health Network, and the Toronto Western Hospital Foundation.

M.H. Li is with the Institute of Biomaterials and Biomedical Engineering, University of Toronto, Toronto, ON, Canada and the Toronto Rehabilitation Institute, University Health Network, Toronto, ON, Canada.

T.A. Mestre was with Toronto Western Hospital, University Health Network, Toronto, ON, Canada. He is now with The Ottawa Hospital Research Institute and Department of Medicine, University of Ottawa Brain and Mind Institute, Ottawa, ON, Canada.

S.H. Fox is with the Division of Neurology, University of Toronto, Toronto, ON, Canada and Toronto Western Hospital, University Health Network, Toronto, ON, Canada.

*B. Taati is with the Toronto Rehabilitation Institute, University Health Network, Toronto, ON, Canada, the Department of Computer Science, University of Toronto, ON, Canada and the Institute of Biomaterials and Biomedical Engineering, University of Toronto, Toronto, ON, Canada (correspondence e-mail: babak.taati@uhn.ca).

## I. INTRODUCTION

Parkinson's disease (PD) is the second most common neurodegenerative disorder after Alzheimer's disease [1], affecting more than 10 million people worldwide [2]. The cardinal features of PD are bradykinesia (slowness of movement), followed by tremor at rest, rigidity, and postural instability [3]. Prevalence of PD increases rapidly over the age of 60 [4], and both global incidence and economic costs associated with PD are expected to rise rapidly in the near future [5], [6]. Since its discovery in the 1970s, levodopa has been the gold standard treatment for PD and is highly effective at improving motor symptoms [7]. However, after prolonged levodopa therapy, 40% of individuals develop levodopa-induced dyskinesia (LID) within 4-6 years [8]. LIDs are involuntary movements characterized by a non-rhythmic motion flowing from one body part to another (chorea) and/or involuntary contractions of opposing muscles causing twisting of the body into abnormal postures (dystonia) [9].

To provide optimal relief of parkinsonism and dyskinesia, treatment regimens must be tailored on an individual basis. While PD patients regularly consult their neurologists to inform treatment adjustments, these consultations occur intermittently and can fail to identify important changes in a patient's condition. Furthermore, the standard clinical rating scales used to record characteristics of PD symptoms require specialized training to perform, and are associated with the inherent subjectivity of the rater [10]. Paper diaries have also been used for patient self-reports of symptoms, but patient compliance is low and interpretation of symptoms can differ significantly between patients and physicians [11], [12].

Computerized assessments are an attractive potential solution, allowing automated evaluation of PD signs to be performed more frequently without the assistance of a clinician. The information gathered from these assessments can be relayed to a neurologist to supplement existing clinic visits and inform changes in management. In addition, computerized assessments are expected to provide an objective measurement of signs, and therefore be more consistent than a patient self-report. Computer vision is an appealing modality for assessment of PD and LID: a vision-based system would be completely noncontact, and require minimal instrumentation in the form of a camera for data capture and computer for processing. The recent emergence of deep learning has produced state-of-the-art performance on many challenging problems, including human pose estimation.

In this paper, deep learning pose estimation algorithms are applied to extract full body movements from videos of specific tasks from clinical PD assessments. Features characterizing the motions are computed from movement trajectories. Afterwards, the features are used to train machine learning algorithms to classify PD/LID motions and estimate involuntary movement severity on clinical rating scales.

Preliminary results from this work have been presented at the Engineering in Medicine and Biology Conference [13]. This paper extends the previous material by analyzing additional motor tasks for parkinsonism as well as a comprehensive evaluation of the predictive power of the chosen feature set.

## II. BACKGROUND

To address the inherent subjectivity and inconvenience of current practices in PD assessment, efforts have been made to develop systems capable of objective evaluation of signs. Studies generally involve the recording of motion signals while participants perform tasks from clinical rating scales, or execute a predefined protocol of activities of daily living (ADL).

Wearable sensing has thus far been the most popular technology for PD assessment, using accelerometers, gyroscopes, and/or magnetometers to record movements. These sensors are often packaged together as inertial movement units (IMU). Keijsers et al. continuously monitored participants during a 35 item ADL protocol and predicted dyskinesia severity in one minute time intervals [14]. Focusing on upper limb movements, Salarian et al. attached gyroscopes to the forearms to estimate tremor and bradykinesia severity [15], while Giuffrida et al. used a custom finger mounted sensor to estimate severity of rest, postural, and kinetic tremors [16]. Patel et al. investigated multiple tasks from the Unified Parkinson's Disease Rating Scale (UPDRS) motor assessment to determine the best tasks and movement features for predicting tremor, bradykinesia, and dyskinesia severity [17]. Delrobaei et al. used a motion capture suit comprised of multiple IMUs to track joint angles and generate a dyskinesia severity score that correlated well with clinical scores [18]. While wearable systems have the potential to be implemented in a discreet and wireless fashion, they still require physical contact with the body. Furthermore, standardization is required regarding the quantity and placement of sensors needed to capture useful movement signals.

In contrast to wearable sensors, vision-based assessment requires only a camera for data capture and computer for processing. These assessments are noncontact, and do not require additional instrumentation to capture more body parts. However, the current state of vision-based assessment for PD and LID is very limited. Multi-colored suits were used for body part segmentation in parkinsonian gait analysis [19], [20], or environments were controlled to simplify extraction of relevant movements [21], [22]. Points on the body were also manually landmarked in video and tracked using image registration to observe global dyskinesia [23]. More complex camera hardware (e.g. Microsoft Kinect) can track motion in 3D with depth sensors and has been used to characterize hand movements [24], as well as analyze parkinsonian gait [25], [26] and assess dyskinesia severity [27] using the Kinect's skeletal tracking capabilities. Multi-camera motion capture systems can capture 3D movements more accurately by tracking the position of reflective markers attached to the points of interest. While they have been explored in the context of PD [28], [29], their prohibitive costs and complicated experimental setup make them impractical outside of research use.

While human pose estimation in video has been actively studied in computer science for several decades, the recent emergence of deep learning has led to substantial improvements in accuracy. Deep learning is a branch of machine learning built on a biologically inspired algorithm called a neural network. Neural networks are made up of layers of neurons that individually perform basic operations, but can be connected and trained to learn complex data representations. One major advantage of deep learning is automatic discovery of useful features, while conventional machine learning approaches use hand engineered features that require domain knowledge to achieve good performance. Convolutional neural networks (CNNs) are a specific deep learning architecture that takes advantage of inherent properties of images to improve efficiency. Toshev and Szegedy were the first to apply deep learning for pose estimation, where they framed joint position prediction as a cascaded regression problem using CNNs as regressors [30]. Chen and Yuille took advantage of the representational power of CNNs to learn the conditional probabilities of the presence of body parts and their spatial relations in a graphical model of pose [31]. Wei et al. iteratively refined joint positions by incorporating long range interactions between body parts over multiple stages of replicated CNNs [32].

The use of deep learning for PD assessment is still in early stages, although a few recent studies have applied deep learning for classification of wearable sensor data [33], [34]. Therefore, an excellent opportunity exists to assess the feasibility of deep learning for vision-based assessment of PD.

## III. METHODS

### A. Dataset

Data was recorded at the Movement Disorders Centre of Toronto Western Hospital with approval from the University Health Network Research Ethics Board and written informed consent from all participants. The primary purpose of the initial study was to determine clinically important changes in parkinsonism and LID rating scales, including the UPDRS and UDysRS (Unified Dyskinesia Rating Scale). Results of the study and detailed information about the protocol are available in [35]. Participants completed a levodopa infusion protocol that allows a standard assessment of PD and LID severity. Assessments were performed at regular intervals using tasks from standard clinical rating scales for parkinsonism and LID.

Videos were captured using a consumer grade video camera at 30 frames per second at a resolution of 480×640 or 540×960. The participants were seated and facing the camera in all videos. All videos were rated by two or three neurologists who were blinded to the time elapsed when the video was recorded.

Nine participants (5 men, median age 64 years) completed the study. All participants had a diagnosis of idiopathic PD and stable bothersome peak-dose LID for more than 25% of the day, defined as a rating ≥ 2 on UPDRS item 4.1 (Time Spent with Dyskinesias) and a rating ≥ 1 on the Lang-Fahn Activities of Daily Living Dyskinesia Scale. The UDysRS Part III was used to rate the severity of dyskinesia and the UPDRS Part III was used to rate the severity of parkinsonism. Due to practical reasons, the dressing task was omitted from the UDysRS and the rigidity assessment was omitted from the UPDRS. While these rating scales have been validated based on clinimetric properties, the single items that comprise the scales have not been validated as standalone measures. A subset of tasks was selected for automated assessment based on perceived feasibility of vision-based analysis and on correlation to the total validated assessment score. The tasks selected were:

- Communication (UDysRS Part III) – the participant describes an image, engages in discussion with the examiner, mental math or recall
- Drinking from a cup (UDysRS Part III)
- Leg agility (UPDRS Part 3.8) – stomping of the leg vertically with as much speed and amplitude as possible
- Toe tapping (UPDRS Part 3.7)

The UDysRS Part III contains seven scores for each task for different parts of the body from 0 (no dyskinesia) to 4 (incapacitating dyskinesia). The seven parts of the body rated are the face, neck, left and right arm/shoulder, left and right leg/hip, and trunk. The total validated score is the sum of the seven highest scores for each body part across all tasks. The UPDRS Part III also uses a four-point scale for severity in each task, and body parts may be rated separately depending on the task. For leg agility and toe tapping, there are ratings for the left and right sides of the body, and these tasks are designed to capture bradykinesia. The total validated score for the UPDRS Part III is the sum of 28 available item scores. The tasks of interest were manually segmented from the complete assessment videos and videos containing severe occlusions or camera movement were removed. Video information can be found in Table I. An anonymized dataset including pose trajectories and clinical scores has been made available.[1]

### B. Trajectory Extraction

Extraction of movement trajectories was conducted using Convolutional Pose Machines (CPM), a state-of-the-art deep learning based pose estimation algorithm [32]. CPM was pre-trained on the MPII Human Pose Dataset, which contained 25 000 images with annotated body joints and covered over 400 human activities [36]. To assist pose estimation, a bounding box was annotated around the participant in the first frame of each video. Video frames were resized and padded to 368×368 before being input to CPM. The output of CPM was a 14-point skeleton with annotation of the head, neck, shoulders, elbows, wrists, hips, knees, and ankles. Joint trajectories were extracted independently for each frame. Sample detections are shown in Fig. 1.

As tasks captured different facets of PD and LID, preprocessing strategies were tailored for each task.

TABLE I
VIDEO DURATIONS FOR EACH TASK

| Task | # of videos | Total duration (h:mm:ss) | Average duration (s) |
|---|---|---|---|
| Communication | 134 | 1:13:26 | 32.9 |
| Drinking | 124 | 15:20 | 7.4 |
| Leg agility | 134 | 24:05 | 10.8 |
| Toe tapping | 134 | 21:17 | 9.5 |

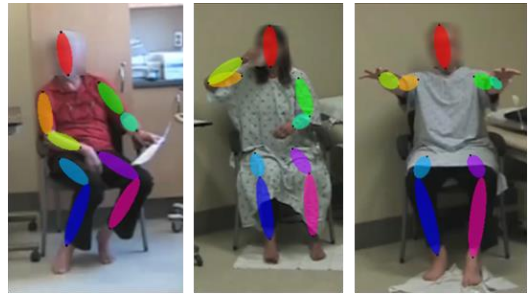

Fig. 1. Sample pose detections produced by CPM.

*1) Communication and Drinking*

Both communication and drinking tasks were rated using the UDysRS Part III, which contains seven subscores for dyskinesia of the face, neck, arms, trunk, and legs. The face dyskinesia subscore was not considered as measurement of facial dyskinesia requires more complex modelling than available through pose estimation.

  a) *Camera shake removal* – Camera motion was isolated by tracking the movement of stationary points in the scene. This was done by detecting and tracking points outside the bounding box where the person was identified using the Kanade-Lucas-Tomasi (KLT) tracker. A maximum of 500 points were tracked, and the median of the frame-to-frame motions was taken as the camera trajectory. Joint trajectories were stabilized by subtracting the camera trajectory.
  b) *Discontinuity removal* – Due to the frame-by-frame nature of the pose estimation approach, temporarily poor estimation can introduce large discontinuities in the joint trajectories. To identify discontinuities, a threshold was placed on the 2D frame-to-frame motion. The threshold was half of the head length, so that the threshold would be invariant to the distance of the participant from the camera. Joint trajectories were split when the trajectory was exceeded, creating multiple segments. The goal of grouping segments is to identify segments that were similarly located spatially and to reject outliers. Grouping of segments proceeded as a forward temporal pass of the entire trajectory. For the current segment, the separation distance between the start of the segment and the end of the existing segment groups was computed. The current segment was added to the group with the minimum separation distance provided the distance was less than the threshold. If this constraint could not be satisfied, the segment became a new group. The confidence of pose estimations from CPM was used to determine which group of segments was most likely to reflect the actual movement. The confidence was the height of the maximum on the heatmap produced by CPM indicating the joint location. The group of segments with the highest median confidence was selected, and gaps between segments were filled using linear interpolation. Segments that did not span the entire signal were truncated at the segment end points.

---
[1] https://github.com/limi44/Parkinson-s-Pose-Estimation-Dataset

  c) *Face tracking* - Although the skeleton from CPM contains a head annotation, it is located on the top of the head and was therefore unsuitable for tracking head turning. To resolve this, a bounding box was placed on the face, which was tracked using the MEEM object tracker [37]. The bounding box was initialized as a square centered at the midpoint between the head and neck annotations, where the side length was the vertical distance between the head and neck. The bottom two thirds and middle 50% horizontally of the square are used as the final bounding box. The bounding box was tracked over time using MEEM and the motion of the center of the bounding box was taken as the face trajectory. The face trajectory replaced the head and neck trajectories from CPM.

*2) Leg agility*

Leg agility bradykinesia was assessed using the UPDRS Part 3.8, containing two item scores for the left and right side. Camera shake removal was the same as for the communication and drinking tasks. However, a low pass filter was applied in lieu of discontinuity removal for smoothing. The filter was a 5th order Butterworth filter with a cut-off frequency of 5 Hz.

*3) Toe tapping*

Toe tapping bradykinesia was assessed using the UPDRS Part 3.7, which contains two item scores for the left and right feet. As the skeleton from CPM included ankle locations and not the feet, dense optical flow was used to capture the toe tapping movements [38]. It was assumed that the heel was anchored to the floor during the task, such that there would not be significant ankle motion. Therefore, the median ankle position in the video was used to infer the area of the foot. A square bounding box was positioned below the ankle, such that the ankle was at the center of the top edge. The side length was the head length and the bounding box was truncated if it extended beyond the video frame.

Given a set of frame-to-frame optical flows, the aggregate toe tapping velocity was computed as the median of non-zero optical flows. Flow velocities greater than $5.0 \times 10^{-4}$ pixels/frame were considered non-zero.

*C. Feature Extraction*

A total of 13 joint trajectories exist after CPM and preprocessing. These trajectories are the left and right shoulders (Lsho, Rsho), elbows (Lelb, Relb), wrists (Lwri, Rwri), hips (Lhip, Rhip), knees (Lkne, Rkne), ankles (Lank, Rank) from the CPM skeleton and the face trajectory from MEEM. Trajectories were normalized by head length to ensure features were comparable across videos. A Savitzky-Golay filter (polynomial order = 3, window length = 11 samples) was used for smoothing and for computing signal derivatives. As each task rating contains subscores that are focused on different anatomical regions, only relevant joint trajectories were used for each subscore. The joints used for each task are shown in Table II.

TABLE II
JOINT TRAJECTORIES FOR EACH TASK

| Task | Subscore | Joints used |
|---|---|---|
| Communication/Drinking (UDysRS) | Neck | Face |
| | Rarm | Rsho, Relb, Rwri |
| | Larm | Lsho, Lelb, Lwri |
| | Trunk | Rsho, Lsho |
| | Rleg | Rhip, Rkne, Rank |
| | Lleg | Lhip, Lkne, Lank |
| Leg agility (UPDRS) | Right | Rhip, Rkne, Rank |
| | Left | Lhip, Lkne, Lank |
| Toe tapping (UPDRS) | Right | Rank* |
| | Left | Lank* |

*For the toe tapping task, ankle locations were used to create a bounding box for motion extraction.

For all tasks besides toe tapping, 32 features were extracted per joint trajectory. There were 15 kinematic features: the maximum, median, mean, standard deviation, and interquartile range of speed, acceleration, and jerk. Spectral features were computed from the Welch power spectral density (PSD) of the displacement and velocity signals. The horizontal and vertical components of the movement signal were combined as a complex signal before spectral estimation to produce an asymmetric spectrum. Afterwards, the positive and negative halves of the full spectrum were summed. There was a total of 16 spectral features: the peak magnitude, entropy, total power, half point (i.e. frequency that divides spectral power into equal halves), and power bands $0.5 - 1$ Hz, $> 2$ Hz, $> 4$ Hz, $> 6$ Hz for both the displacement and velocity PSDs. The PSDs were normalized before computing power bands such that they were relative to the total power.

Since the signal for the toe tapping task was an aggregate velocity, the feature extraction approach was modified. Kinematic features were computed separately for the total speed and for the horizontal and vertical velocities. In addition to the 15 features used for the other tasks, skew and kurtosis were also computed for speed, acceleration, and jerk, yielding 21 features per signal for a total of 63 kinematic features. As there was no displacement signal, spectral features were only extracted from the velocity signal. The horizontal and vertical components of the aggregate velocity were used to compute four velocity PSDs: combined horizontal and vertical as a complex signal, horizontal only, vertical only, and magnitude of velocity. Each PSD had eight features, for a total of 32 spectral features. Convex hull could not be computed without a displacement signal. Overall, there were 95 features per joint for the toe tapping task.

As the communication task involved multiple subtasks, transitions between subtasks often contained voluntary movements or the video was cut by the examiner. Therefore, the communication task was divided into subtasks, features were computed for each component and then averaged to get the overall communication task features.

*D. Evaluation*

All experiments were performed using leave-one-subject-out cross-validation and random forest. Specific implementation details and metrics are described in the following sections. Random forest hyperparameters were selected using 200 iterations of randomized search. Possible values for hyperparameters are given in Table III ($m$ = number of features). All intervals are integer intervals.

*1) Binary classification*

Binary classification can be framed as the detection of pathological motion, whether PD or LID. For each subscore of the UDysRS and UPDRS, ratings were on a scale of 0-4, where 0 indicated normal motion and 4 indicated severe impairment. Ratings from neurologists were averaged. For the communication and drinking tasks, a threshold of 0.5 was used for binarizing scores, where average scores equal to or less than 0.5 were considered normal motion. For the leg agility and toe tapping tasks, there were fewer low ratings so thresholds of 1 and less than 2 (not inclusive) were selected, respectively, for binarization of scores to balance classes. Metrics used were the F1-score and area under the curve (AUC).

*2) Regression*

The goal of regression is prediction of the clinical rating of PD or LID severity based on movement features. Ratings were predicted for each task individually, and features were combined to predict the total validated scores of the UDysRS Part III and UPDRS Part III. The total validated score for the UDysRS Part III contains the highest subscores for each body part across all tasks (0-4) and the sum of subscores (0-28), while the total validated score for the UPDRS Part III was the sum of all task scores (0-112). For the UDysRS Part III, features were combined from the communication and drinking tasks. For the UPDRS Part III, features were combined from the communication, leg agility, and toe tapping tasks. The leg agility task features include not only those in Table II but for all recorded joints. Metrics used were the RMS error and Pearson correlation between predictions and clinician ratings.

*3) Multi-class classification*

There are three possible classifications of motions – PD, PD with LID, or normal. For tasks to be suitable, they require ratings for both PD and LID. Although the communication task does not explicitly have a rating for PD, the UPDRS Part 3.14 (Global spontaneity of movement) is used as a replacement as it is a global rating of PD. Ratings are averaged across all applicable body part subscores to generate a single severity score. Given ratings of both PD and LID, if neither score was greater than 1, the motion was considered normal. Otherwise, the motion was assigned the label corresponding to the higher score. If ratings were equal, the motion was omitted from further analysis. The metric used to assess performance was accuracy.

## IV. RESULTS

Binary classification and regression results for communication and drinking tasks are shown in Table IV, while results for the leg agility and toe tapping tasks are given in Table V. Mean correlations were computed using Fisher z-transformation. For binary classification, the number of ratings binarized to the negative class is denoted by $n_0$ and informs if the classification task was well balanced. There are some disparities between the number of videos (Table I) and the number of samples shown in Table IV and Table V, as some

TABLE III
HYPERPARAMETERS FOR RANDOM FOREST

| Hyperparameter | Possible values | |
|---|---|---|
| | Classification (Binary/Multiclass) | Regression |
| Max features to try | $[1, ..., \lfloor\sqrt{m}\rfloor]$ | $[1, ..., \lfloor m/3 \rfloor]$ |
| Min samples to split node | $[1, ..., 11]$ | |
| Min samples to be leaf node | $[1, ..., 11]$ | |
| Number of trees | $[25, ..., 50]$* | |
| Impurity criterion | Gini index/Entropy | N/A |

*except UPDRS Part III total score, $[64, ..., 128]$

TABLE IV
RESULTS FOR COMMUNICATION AND DRINKING TASKS

| | Communication ($n = 128$) | | | | | | | Drinking ($n = 118$) | | | | | | |
|---|---|---|---|---|---|---|---|---|---|---|---|---|---|---|
| Regression | Neck | Rarm | Larm | Trunk | Rleg | Lleg | Mean | Neck | Rarm | Larm | Trunk | Rleg | Lleg | Mean |
| RMS | 0.559 | 0.399 | 0.465 | 0.513 | 0.579 | 0.590 | 0.518 | 0.724 | 0.737 | 0.575 | 0.701 | 0.586 | 0.622 | 0.657 |
| $r$ | 0.712 | 0.760 | 0.645 | 0.760 | 0.522 | 0.490 | 0.661 | 0.075 | -0.150 | -0.003 | 0.099 | 0.087 | 0.147 | 0.043 |
| Binary Classification | Neck $n_0 = 48$ | Rarm $n_0 = 60$ | Larm $n_0 = 54$ | Trunk $n_0 = 60$ | Rleg $n_0 = 57$ | Lleg $n_0 = 59$ | Mean | Neck $n_0 = 61$ | Rarm $n_0 = 79$ | Larm $n_0 = 81$ | Trunk $n_0 = 60$ | Rleg $n_0 = 70$ | Lleg $n_0 = 66$ | Mean |
| F1 | 0.941 | 0.920 | 0.929 | 0.960 | 0.819 | 0.865 | 0.906 | 0.711 | 0.148 | 0.289 | 0.643 | 0.594 | 0.617 | 0.500 |
| AUC | 0.935 | 0.957 | 0.946 | 0.983 | 0.852 | 0.907 | 0.930 | 0.774 | 0.418 | 0.557 | 0.687 | 0.673 | 0.696 | 0.634 |

videos did not have all possible ratings available.

The mean correlation between LID severity predictions and ground truth ratings for the communication task was 0.661, compared to 0.043 for the drinking task. For PD severity predictions, the mean correlations were 0.618 and 0.372 for the leg agility and toe tapping tasks, respectively. Binary classification of communication task features achieved a mean AUC of 0.930, while drinking task performance had a mean AUC of 0.634. For the leg agility task, the mean AUC was 0.770, while the AUC for the toe tapping task was 0.773.

TABLE V
RESULTS FOR LEG AGILITY AND TOE TAPPING TASKS

|  | Leg agility ($n = 75$) | | | Toe tapping ($n = 76$) | | |
| --- | --- | --- | --- | --- | --- | --- |
| Regression | Right | Left | Mean | Right | Left | Mean |
| RMS | 0.648 | 0.462 | 0.555 | 0.614 | 0.615 | 0.614 |
| $r$ | 0.504 | 0.710 | 0.618 | 0.383 | 0.360 | 0.372 |
| Binary Classification | Right $n_0 = 43$ | Left $n_0 = 36$ | Mean | Right $n_0 = 39$ | Left $n_0 = 36$ | Mean |
| F1 | 0.538 | 0.725 | 0.631 | 0.755 | 0.694 | 0.725 |
| AUC | 0.699 | 0.842 | 0.770 | 0.842 | 0.704 | 0.773 |

For multiclass classification, the overall accuracy on the communication task was 71.4%. Sensitivity and specificity for each class are provided in Table VI. For predicting the total validated scores on the UDysRS Part III and UPDRS Part III, the results are given in Table VII. The correlation between predicted and ground truth ratings was 0.741 and 0.530 for the UDysRS and UPDRS, respectively.

TABLE VI
MULTICLASS CLASSIFICATION RESULTS FOR COMMUNICATION TASK

|  | $n$ | Sensitivity | Specificity |
| --- | --- | --- | --- |
| LID | 26 | 96.2% | 95.7% |
| Normal | 17 | 9.4% | 89.7% |
| PD | 34 | 83.5% | 68.4% |
| Overall Accuracy | 77 | 71.4% | |

TABLE VII
RESULTS FOR PREDICTION OF VALIDATED SCORES

| Regression | UDysRS Part III ($n = 118$) | UPDRS Part III ($n = 74$) |
| --- | --- | --- |
| RMS | 2.906 | 7.765 |
| $r$ | 0.741* | 0.530* |

## V. DISCUSSION

The task with the best performance was the communication task. This was not surprising, as it is well-known clinically that the communication task elicits involuntary movements [39]. In contrast, performance on the drinking task was poor, despite the RMS error appearing similar. This was because most ratings for the drinking task were between 0 and 2, thus emphasizing that both RMS and correlation are necessary to accurately portray performance. However, features from the drinking task still had discriminative power for detecting dyskinesia, as the mean AUC was better than 0.5. Drinking task arm subscore performance was noticeably worse than for other subscores, which was likely due to inability to discern voluntary from involuntary movements, as well as increased occlusion of upper limbs during movement. For multiclass classification of the communication task, the class that was best discriminated was LID, while many of the normal class were incorrectly classified as PD. Intuitively, the communication task does not prompt participants to move voluntarily, therefore the slowness or absence of movement in PD and the lack of voluntary movement in the normal class can be confused with each other. This contrasts with the larger involuntary movements present in LID, which are easily identifiable.

Although features from a subset of tasks are used to predict the total UPDRS Part III and UDysRS Part III scores, predictions had moderate to good correlation with total scores. This indicates that using this technology, an abbreviated version of these clinical scales could be sufficient. Previous studies have used measures derived from simple tasks such as the timed up and go [40] and a touchscreen finger tapping and spiral drawing test [41] to achieve moderate to good correlation with the total UPDRS Part III score.

No explicit feature selection was performed despite having a large number of features compared to samples. Although the random forest algorithm is generally resistant to overfitting, feature selection can often still reduce features that are not useful. However, after evaluating several feature selection methods, no performance boost was observed compared to applying random forest with all features. Dimensionality reduction methods were not tested as feature transformation would reduce interpretability, thus making further analysis more difficult. Likewise, more complex algorithms that learn feature representations were not considered as discovered features may not have been clinically useful.

While the results indicate that CPM could capture clinically relevant features from videos, this serves as an indirect measure of the accuracy of pose estimation. In preliminary testing, a benchmark made of frames of video from the dataset was used to assess CPM. All body parts were well detected except for the knees. Knee detection was complicated due to the hospital gowns worn by participants, which resulted in insufficient texture to discern knee location. This means that the involuntary opening and closing motions of the knees were poorly tracked, which may explain why leg subscore predictions were the worst in the communication task. However, ankles were well tracked so this is not expected to have significantly affected performance on the leg agility task.

Due to differing experimental conditions and rating scales used in past studies, it is difficult to perform a direct comparison in terms of system performance. The closest study in terms of experimental protocol was Rao et al., who analyzed videos of the communication task and tracked manually landmarked joint locations to develop a dyskinesia severity score [23]. They report good correlation between their score and the UDysRS Part IV (single rating of disability) score (Kendall tau-b correlation 0.68 – 0.85 for different neurologists). Their study used non-rigid image registration for tracking, which was not able to infer joint positions if occluded, and could not recover if the joint position was lost. In contrast, deep learning based pose estimation learns the structure of the human body after seeing training data, and can often make accurate predictions of joint locations even when the joints are not visible. Dyshel et al. leveraged the Kinect's skeletal tracking to extract movement parameters from tasks from the UPDRS and AIMS (Abnormal Involuntary Movement Scale) [27]. They trained a classifier to detect dyskinesia with an AUC of 0.906, and quantified the dyskinesia severity based on the percent of a movement classified as dyskinetic. This quantitative measure had good correlation with AIMS scores (general correlation coefficient 0.805). In wearable sensing, Patel et al. reported classification errors of 1.7% and 1.2% for bradykinesia and dyskinesia, respectively, using tasks from the UPDRS [17]. Tsipouras et al. detected dyskinesia with 92.51% accuracy in a continuous recording of multiple ADLs [42]. Eskofier et al. used CNNs on accelerometer recordings of the pronation/supination and hand movements tasks and achieved bradykinesia classification accuracy of 90.9% [33]. In this work, the best performance for binary classification of dyskinesia was in the communication task, with an AUC of 0.930. This is comparable with other studies, including those using wearables, although the difficulty of classification is highly dependent on the length of the motion segments to be classified and the type of motion performed. For parkinsonism/bradykinesia, the best binary classification performance was for the toe tapping task, with an AUC of 0.773. This is not as high as dyskinesia classification performance, and can likely be attributed to the distribution of ratings. In the communication task, 30-40% of ratings for subscores were at the lower limit of the scale (i.e. 0), whereas for the leg agility and toe tapping tasks, this percentage was much smaller (less than 3%). Threshold selection for binarizing scores was based on balancing classes, and therefore may not have been optimal with respect to clinical definitions.

## VI. LIMITATIONS

As the videos from this dataset were not captured for subsequent computer vision analysis, there were recording issues that introduced noise, including different camera angles and zoom. Videos were also recorded in 2D, resulting in loss of depth information; this made it difficult to resolve joint angles that may help with detection of LID. Despite these concerns, the videos are representative of the quality of videos used by clinicians for PD assessment, and the availability of the data outweighed the unnecessary burden on participants required to perform a new experiment. However, manual intervention was required for task segmentation and person localization. For this feasibility study, the videos were of sufficient quality and future refinement of recording protocols should improve algorithm performance and consistency. Future studies could use deep learning algorithms that take advantage of temporal information in videos for more accurate pose estimation [43].

The recruitment criteria selected individuals with moderate levels of dyskinesia. Therefore, the study population reflects only a segment of the patient population. In addition, a small number of tasks from the UPDRS and UDysRS were not assessed for practical reasons. While adjustments of rating scales are common practice, clinimetrics have not been tested on subsets of the complete evaluation and can potentially compromise validity. The selection of thresholds for binarization was directed more by a desire to balance classes than by the clinical definitions of those thresholds. Ideally, the solution would be to gather sufficient data to represent all ratings and to select thresholds either based on clinical supervision or by discovery of an optimal separation between groups.

Regression performance is reported using correlation; however, it is unclear what would be a clinically useful level of agreement. Furthermore, while a high correlation may indicate that a method is able to mimic clinicians, validation based on agreement with clinical ratings does not provide insight into whether such technologies can achieve better sensitivity to clinically important changes than subjective rating scales. Additional investigation is required to compare the sensitivity of the proposed system to validated clinical measures.

## VII. CONCLUSION

This paper presents the first application of deep learning for vision-based assessment of parkinsonism and LID. The results demonstrate that state-of-the-art pose estimation algorithms can extract meaningful information about PD motor signs from videos of Parkinson's assessments and provide a performance baseline for future studies of PD with deep learning. The long-term goal for this system is deployment in a mobile or tablet application. For home usage, the application could be used by patients to perform regular self-assessments and relay the information to their doctor to provide objective supplemental information for their next clinic visit. An automated system capable of detecting changes in symptom severity could also have major impact in accelerating clinical trials for new therapies.